\documentclass{article}

\usepackage{microtype}
\usepackage{graphicx}
\usepackage{subfigure}
\usepackage{booktabs} % for professional tables

\usepackage{hyperref}

\usepackage[accepted]{icml2024}

\usepackage{amsmath}
\usepackage{amssymb}
\usepackage{mathtools}
\usepackage{amsthm}
\usepackage{nccmath}% for flush left equations: \begin{fleqn} \end{fleqn}

\usepackage[capitalize,noabbrev]{cleveref}

\theoremstyle{plain}

\theoremstyle{definition}

\theoremstyle{remark}

\usepackage[textsize=tiny]{todonotes}

\usepackage{xcolor}
\definecolor{C}{HTML}{0047AB}
\hypersetup{
  colorlinks   = true, %Colours links instead of ugly boxes
  urlcolor     = C, %Colour for external hyperlinks
 } 

\icmltitlerunning{Baba Is AI: Break the Rules to Beat the Benchmark}

\begin{document}

\twocolumn[
\icmltitle{Baba Is AI: Break the Rules to Beat the Benchmark}



% \icmlsetsymbol{equal}{\textbf{*}}
\icmlsetsymbol{equal}{$\boldsymbol{*}$}

\begin{icmlauthorlist}
\icmlauthor{Nathan Cloos}{MIT}
\icmlauthor{Meagan Jens}{MIT}
\icmlauthor{Michelangelo Naim}{MIT}
\icmlauthor{Yen-Ling Kuo}{Vir}
\icmlauthor{Ignacio Cases}{MIT}\\
% \icmlauthor{Andrei Barbu$^\boldsymbol{*}$}{MIT}
% \icmlauthor{Christopher J. Cueva$^\boldsymbol{*}$}{MIT}
\icmlauthor{Andrei Barbu*}{MIT}
\icmlauthor{Christopher J. Cueva*}{MIT}
\end{icmlauthorlist}

\icmlaffiliation{MIT}{MIT}
\icmlaffiliation{Vir}{Department of Computer Science, University of Virginia, USA}

\icmlcorrespondingauthor{Nathan Cloos}{nacloos@mit.edu}
\icmlcorrespondingauthor{Andrei Barbu}{abarbu@mit.edu}
\icmlcorrespondingauthor{Christopher J. Cueva}{ccueva@gmail.com}

\icmlkeywords{large language model, grounded compositional generalization, benchmark, baba is you}

\vskip 0.3in
]



\printAffiliationsAndNotice{\icmlEqualContribution} % otherwise use the standard text.

\begin{abstract}
Humans solve problems by following existing rules and procedures, and also by leaps of creativity to redefine those rules and objectives. To probe these abilities, we developed a new benchmark based on the game Baba Is You where an agent manipulates both objects in the environment and rules, represented by movable tiles with words written on them, to reach a specified goal and win the game. We test three state-of-the-art multi-modal large language models (OpenAI GPT-4o, Google Gemini-1.5-Pro and Gemini-1.5-Flash) and find that they fail dramatically when generalization requires that the rules of the game must be manipulated and combined. %systematic composition of previously learned rules.
% CJC 9.8.2025: added link to code
\\

\vspace{-0.1in}
% \href{https://github.com/nacloos/baba-is-ai}{\normalsize \textbf{{Code}}}
\textbf{Code: } \href{https://github.com/nacloos/baba-is-ai}{github.com/nacloos/baba-is-ai}

\vspace{-0.26in}
\end{abstract}

\section{Introduction}
\vspace{-0.05in}% CJC 9.8.2025 decreased vertical space to preserve original text in first column

\begin{figure}[ht]
\vskip 0in
\begin{center}
\centerline{\includegraphics[width=0.8\columnwidth]{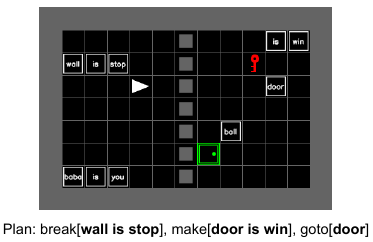}}
\vskip -0.1in
\caption{\textbf{Environment based on the puzzle game Baba Is You.}}
\label{fig:1}
\end{center}
\vskip -0.25in
\end{figure}

Humans demonstrate remarkable abilities in rapid learning and adaptive behavior when faced with novel environments - not only learning and following rules dictated by the environment but altering these rules to enable new outcomes. These abilities leverage two key components that we explore in this paper: 

1) The ability to identify and manipulate relevant stimuli in the environment while ignoring distractor objects and rules. 

2) The ability to combine previously seen rules in novel ways. 

% The ability to use a previously learned rule with new objects.
The ability to study how an agent explicitly learns rules, composes them, and crucially, makes or breaks these rules to alter how the environment and agent behaves, prompted us to develop a new benchmark environment based on the puzzle game Baba Is You. In this game, the player often controls a character named ``Baba" and must navigate through the grid-based world filled with blocks, objects, and textual rules. We can think of this game as a dynamic environment where the player interacts with various objects and rules to achieve specific goals. A remarkable aspect of Baba Is You is that the rules of the game can be manipulated and rearranged by the player.

\begin{figure}[ht]
\vskip 0in
\begin{center}
\centerline{\includegraphics[width=\columnwidth]{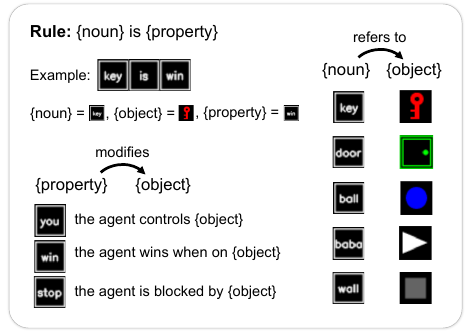}}
\vskip -0.13in
\caption{\textbf{Active rules in the environment modify the properties of the objects.} A rule is active when it is horizontally aligned and has the form $\{$noun$\}$ is $\{$property$\}$.}%A rule is active a noun, a ``is", and a property rule block are horizontally aligned. Each noun refers to a specific type of object in the environment. The property in an active rule modifies the dynamics of the object referred to by the noun.}
\label{fig:2}
\end{center}
\vskip -0.25in
\end{figure}

\begin{figure*}[h]
\vskip 0in
\begin{center}
\centerline{\includegraphics[width=\textwidth]{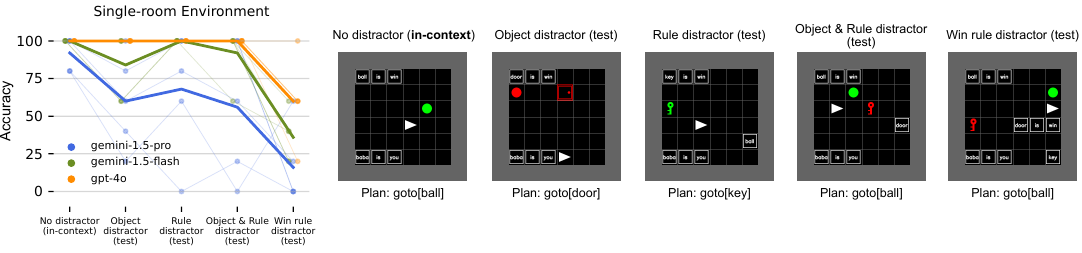}}
\vskip -0.13in
\caption{\textbf{Accuracy of LLMs across 5 environments testing the ability to generalize in the presence of distractors.} The task is to go to the winning object specified by the text box in the active win rule. Accuracy drops substantially on the final task where both an object and an active rule distractor are present. In this final task the irrelevant win rule does not refer to any of the objects in the environment.}
\label{fig:3}
\end{center}
\vskip -0.31in
\end{figure*}

Figure \ref{fig:1} shows an example game environment. The text blocks [baba is you] indicate the player is controlling the white triangle, i.e. the [baba] object, and can now move this object through the environment. Now let's look for the text blocks that specify how to win the game. The [is win] text blocks in the upper right of the environment are incomplete and so the agent must recognize that there is currently no way to win the game until the winning condition is specified. This is accomplished by moving one of the available text block such as [door] or [ball]  to create a rule for winning the game. With this specific environmental layout, a winning strategy is to push the [door] block to create the rule, [door is win], and then move the agent onto the door block, shown in green, to win the game. However, the text blocks [wall is stop] are aligned and so this rule is active and the player cannot move baba through the vertical wall of gray squares to carry out this plan. The player must first push one of the blocks in this rule out of alignment to deactivate the rule [wall is stop]. The final plan to win the game is to first break the rule [wall is stop], then make the rule [door is win], and finally move onto the door object. 

As this example illustrates, this is a dynamic environment where the agent must identify the relevant objects and rules in the environment and then manipulate the environment to change or create rules for success (Figure \ref{fig:2}). 
% We developed the Baba is You environment using OpenAI's minigrid engine \textcolor{red}{references}.
We implemented a simplified version of Baba Is You (Baba Is AI) based on the Gymnasium Minigrid environment \cite{MinigridMiniworld23}.

The goal of the Baba Is AI benchmark is to evaluate the role of systematic compositionality in rule-based generalization. The core component of this benchmark is that the written commands are not only grounded in an environment, but the grounding itself can be manipulated via changing the rules of the environment. This dynamic design allows us to explore a broader notion of generalization compared to the current benchmarks. 

We show results for three large language models (LLMs): GPT-4o, Gemini-1.5-Pro (May 2024), and Gemini-1.5-Flash (May 2024) \cite{geminiteam2024gemini}. We chose GPT-4o and Gemini-1.5-Pro as these models occupy the top two spots on the Chatbot Arena Leaderboard (May 2024) \cite{Chiang2024ChatbotAA}. We also include Gemini-1.5-Flash as this model occupies an intriguing spot in the LLM ecosystem with both excellent performance and affordable price, making it an attractive option for many applications. Previous work often convert visual inputs into text before evaluating LLMs \cite{yao2023react, pmlr-v202-carta23a, ida2023planning, valmeekam2023planbench}. Here we leverage the multi-modal ability of these models to evaluate them directly on visual inputs of the game.

\section{Method}

We first prompt LLMs with general text instructions to play the game. This includes a description of the possible objects and textual rule blocks in the environment, and how active rules can change object properties (as illustrated in Figure \ref{fig:2}, with the exact prompt in Appendix \ref{appendix:1}). Importantly, we specify that a rule is active only if it follows the form ``object is property'' and that the three rule blocks must be aligned horizontally in the environment.

Following previous work on LLM-based agents and planners \cite{ichter2022do, huang2022language, simulacra, valmeekam2023on, song2023, wang2024voyager}, we ask LLMs to operate at a higher level than the low-level control of actions in the environment. Specifically, we ask LLMs to produce high-level textual plans consisting of the following primitives: breaking an active rule, making a rule active, or moving to a specific object in the environment (see an example plan in Figure \ref{fig:1}). We instruct LLMs that these actions can only be taken if the relevant objects and rule blocks are present in the current environment. To generate their plan, LLMs receive as visual input a static image of the initial configuration of the environment.
% These action sequences can serve as plans to guide low-level controllers, even though here we focus on evaluating the planning abilities of the LLM component. 

\begin{figure*}[h]
\begin{center}
\centerline{\includegraphics[width=\textwidth]{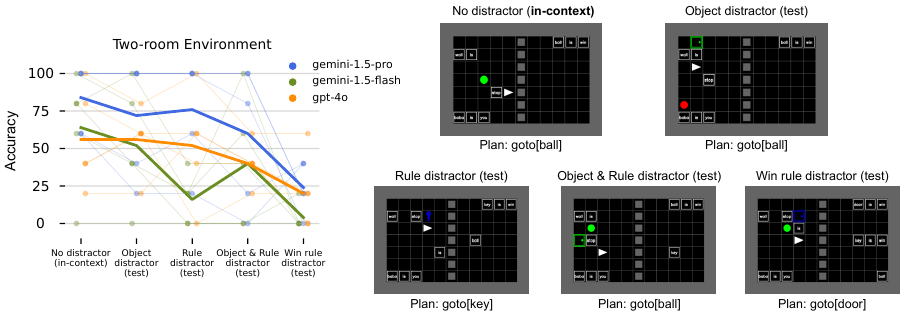}}
\vskip -0.13in
\caption{\textbf{The mean accuracy for all three models is lower when asked to generalize to distractors in a more complex environment.} This environment introduces a central vertical wall. However, the rule [wall is stop] is initially always inactive and so the wall has no practical impact on the movement of the agent. The task is to go to the object referred to by the active win rule (same as in Figure \ref{fig:3}).}
\label{fig:4}
\end{center}
\vskip -0.3in
\end{figure*}

After providing the game instructions, we present LLMs with 10 example images and corresponding winning plans for in-context learning \cite{Brown2020}. 
For each example, LLMs are asked to generate reasoning steps to derive the target plan from the given image. Following the in-context examples, LLMs are prompted to describe a general algorithm to solve the environments and to apply it to unseen test environments. The test environments are specifically chosen to assess different type of generalization. We measure accuracy as the exact match between the final response of  LLMs and the winning plan of the test environment. LLMs are evaluated on 5 samples for each test environment. 

This entire process is repeated for 5 random seeds, each corresponding to different in-context and test examples.

\section{Results}

Our first tests assess the LLMs' ability to extract the most basic rule of the game from in-context examples, namely, go to the winning object, and then apply this rule in novel environments where distractors are present. Complex environments contain not only relevant stimuli but also irrelevant objects or rules; identifying the relevant from irrelevant is a crucial ability that we probe in this set of experiments. 

Figure \ref{fig:3} shows the accuracy of the LLMs in five different environments: 1) Environments without a distractor, i.e. new random variations of the environment used during in-context learning. 2) Environments where there are now two objects but one of them is a distractor. In order to win the game, the agent must go to the object specified in the text box with the win rule, e.g. [door is win] requires the agent to go to the door. 3) Environments contain a noun block that is distracting from the active win rule. 4) Environments contain both a distractor object and noun block. 5) Environments contain both a distractor object and a noun block that is part of an active rule. The distractor rule is not relevant for the environment and so should be ignored. For example, the rightmost panel in Figure \ref{fig:3} shows the distractor rule [door is win] but there is no door object in the environment and so the winning strategy is to follow the other rule [ball is win] and navigate to the ball.

Impressively, GPT-4o performs with perfect accuracy on the first four environments, and as a reminder, this is while receiving visual and not textual inputs about the game. Surprisingly, Gemini-1.5-Flash outperforms Gemini-1.5-Pro, with all models showing the same trend downwards in accuracy on the final task that includes both an object and a rule distractor.

The sequence of environments used to test the LLMs in Figure 4 includes the same distractors as in Figure 3, but now all the environments include a gray vertical wall that runs down the center of the environment. The environments are always initialized with the rule [wall is stop] inactive, as the three blocks that form this rule are not horizontally aligned, and so the wall has no practical impact on the movement of the agent. However, these environments now all contain the extra distractor blocks that compose the inactive wall, and blocks about the wall rule. The mean accuracy for all three models is lower under this increased distractor load (compare Figures 3 and 4). 

Compositional generalization has been studied in many contexts \cite{Lake_Ullman_Tenenbaum_Gershman_2017, Lake2017GeneralizationWS, ruis2020}, for example, if an agent has learned to solve a task with red circles and green keys then it should generalize to red keys and green circles. In the Baba Is AI environment we can not only study these traditional forms of generalization but probe models under scenarios where the very rules of the game must be manipulated and combined. Figure 5 shows one example scenario where the LLMs are shown environments that each highlight three winning strategies and then are asked to solve a new set of environments that require a novel composition of these previously learned rules.

\begin{figure*}[h]
\begin{center}
\centerline{\includegraphics[width=0.9\textwidth]{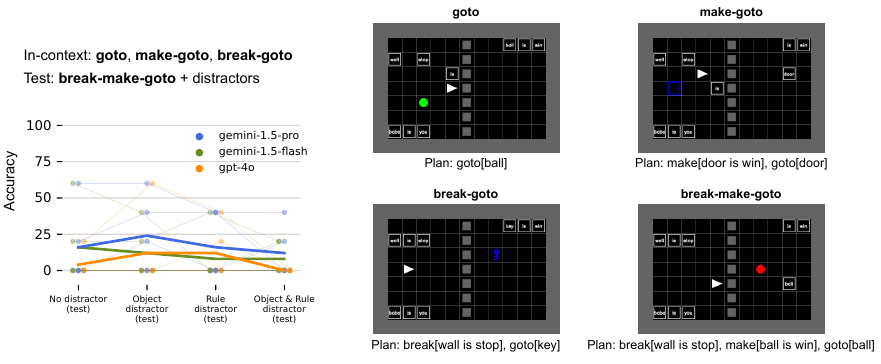}}
\vskip -0.13in
\caption{\textbf{LLMs generalize poorly under scenarios where the rules of the game must be manipulated and combined.} LLMs are shown environments that each highlight three winning strategies: goto$\{$object$\}$; make$\{$rule$\}$ then goto$\{$object$\}$; break$\{$rule$\}$ then goto$\{$object$\}$. Then they are asked to solve a new set of environments that require a novel composition of these previously learned rules: break$\{$rule$\}$ then make$\{$rule$\}$ then goto$\{$object$\}$.}
\label{fig:5}
\end{center}
\vskip -0.34in
\end{figure*}

\begin{figure*}[h]
\vskip 0.1in
\begin{center}
\centerline{\includegraphics[width=0.9\textwidth]{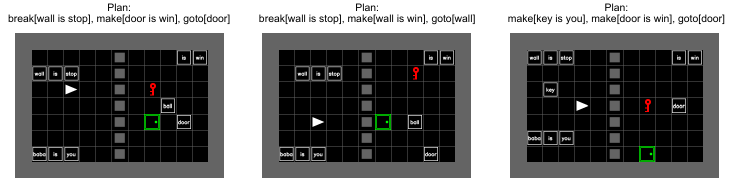}}
\vskip -0.15in
\caption{\textbf{Despite superficial similarities and identical objects, each environment requires distinct winning solutions, illustrating further challenges in rule manipulation and compositional generalization.}}
\label{fig:creative}
\end{center}
\vskip -0.25in
\end{figure*}

%\begin{fleqn}
%\begin{align*}

% falign causes some compilation errors (can't submit to arxiv)
% \begin{falign*}
% \textbf{In-context:}
% \left\{
% \begin{array}{l}
% \text{goto$\{$object$\}$} \\
% \text{make$\{$rule$\}$, goto$\{$object$\}$} \\
% \text{break$\{$rule$\}$, goto$\{$object$\}$}
% \end{array}
% \right.
% \end{falign*}

%\end{align*}
%\end{fleqn}

\vspace{-1em}
\begin{flalign*} \hspace{-6em}
\textbf{In-context:}
\left\{
\begin{array}{l}
\text{goto$\{$object$\}$} \\
\text{make$\{$rule$\}$, goto$\{$object$\}$} \\
\text{break$\{$rule$\}$, goto$\{$object$\}$}
\end{array}
\right.
\end{flalign*}
\vspace{-1em}

\hspace{0.92cm}\textbf{Test:} \hspace{0.4cm} break$\{$rule$\}$, make$\{$rule$\}$, goto$\{$object$\}$
 
% (go-to$\{$object$\}$; make$\{$rule$\}$ then go-to$\{$object$\}$; break$\{$rule$\}$ then go-to$\{$object$\}$), 
% and then are asked to solve a new set of environments that require a novel composition of these previously learned rules: break$\{$rule$\}$ then make$\{$rule$\}$ then go-to$\{$object$\}$.
The accuracy for all three LLMs is low. We have also alternated the four strategies shown in Figure \ref{fig:5} so a different three are used for in-context training and the remaining is used for testing (not shown), and accuracy remains low. 

These aspects of compositional generalization across rules are particularly unique to the Baba Is AI benchmark, and the poor performance indicates that this benchmark creates meaningful generalization challenges for LLMs.

% \section{Related work}
% Keke is AI Competition \cite{charity2022keke}.
% Note that they cite Baba Is You as (Hempuli, 2019) (but no corresponding reference).
% From Keke is AI paper:
% "Baba is You is a puzzle game developed by Arvi “Hempuli” Teikari originally for the 2017 Nordic Game Jam and then expanded to a full game with more mechanics

% Embodied LLM agents. Voyager in Minecraft.
% Here similar because we assume there is a high-level action API and the LLM is prompted to generated commands using this API, instead of directly producing low-level control actions.
% Different because we evaluate how well LLMs can generalize to environments that differ from the in-context examples.

% Evaluate models on text version of minigrid \cite{pmlr-v202-carta23a}
% Here not just text, test LLMs in visually grounded environments

% Recent work has shown that language models can be used as a high level planner for agents interacting in an environment. Given a library of skills a language model proposes a composition of skills to achieve a given goal (e.g. in Minecraft \cite{wang2024voyager}). However, they often rely on textual descriptions
% Assume we have an action api for baba is you with actions make rule, break rule, goto object.
% Is the LLM able to produce appropriate sequences of api actions to solve a given environment solely based on visual input?

% Previous work often convert visual inputs into text before evaluating LLMs. 
% Here we leverage the multi-modal ability of these models to evaluate them directly on visual inputs of the game.

\begin{table}[h]
\centering
\begin{center}
\begin{small}
% \begin{sc}
\begin{tabular}{lc}
\hline
\textbf{Model} & \textbf{Accuracy (mean ± std)} \\
\hline
gemini-1.5-flash & 20.0 ± 29.28 \\
gemini-1.5-pro & 14.67 ± 20.66 \\
gpt-4o & 17.33 ± 28.15 \\
\hline
\end{tabular}
% \end{sc}
\end{small}
\end{center}
\vskip 0in
\caption{Model accuracies for the environments shown in Figure \ref{fig:creative}.}
\label{table:1}
\vskip -0.22in
\end{table}

\section{Discussion}
In order for agents to have human-like interactions with the world, they should not only be able to interact with objects but also have the capacity to understand and manipulate the rules of their environment. By defining a static set of rules that an agent must follow, many games and benchmarks have overlooked a critical capability: the ability to understand rules via rule manipulation. Therefore, the Baba Is AI benchmark explores compositional generalization under conditions in which agents can modify the rules of the environment. Figure \ref{fig:creative} illustrates some of the further challenges in these environments. All three environments are superficially similar and contain the same objects, yet the winning solutions are different in each case (see text at the top of the figures). For example, the center environment requires the agent to break the [wall is stop] rule, then move the [wall] block to create the rule [wall is win], and finally go to one of the wall blocks to win the game. As a second example, in the environment shown in the rightmost panel of Figure \ref{fig:creative} the rule [wall is stop] is located in the corner of the environment and so there is no way to push these blocks out of alignment and break this rule; the agent is initially trapped in the leftmost room of the environment. The agent must break the currently active rule [baba is you] and create [key is you] in order to control the key on the other side of the wall. Then the agent can use the key to create the rule [door is win] and move to the door. The accuracy on these challenging environments is low as shown in Table \ref{table:1}. 

The errors that LLMs make in solving the Baba Is AI environments are instructive about future opportunities for improvements (see Appendix \ref{appendix:2}). LLMs make grounding mistakes: the LLM refers to an object that does not exist in the environment. LLMs make path planning mistakes: the LLM incorrectly asserts that the path to a specific object is blocked by another object, despite the path being clear in the environment.

%Common error cases: fail to locate object spatially
%Direction to improve the models: give them access to a spatial affordance module

% Models are not able to interact with the environment.
% Humans rapidly learn to solve new environments with exploration

% When humans encounter novel environments, they rapidly leverage a wealth of priors to understand the goal, identify the agent they control, and determine how to manipulate the agent to achieve the goal \cite{dubey2018investigating}. 
% This capability extends even to environments never seen before, aided by our visual priors—such as associating red with danger and green with safety—and the ability to quickly infer control dynamics through minimal interactions. 
% For instance, pressing keys to observe resulting movements allows us to identify the controllable agent in an environment \citep{Tenenbaum_Ullman_2023}. Additionally, humans can rapidly learn the rules of a game from a few interactions and explore systematically to determine the type of various objects such as enemy and goal objects\citep{tsividis2021humanlevel}.

% Here we have only scratched the surface of the possibilities that Baba Is You offer.

\clearpage

\bibliography{example_paper}
\bibliographystyle{icml2024}

%%%%%%%%%%%%%%%%%%%%%%%%%%%%%%%%%%%%%%%%%%%%%%%%%%%%%%%%%%%%%%%%%%%%%%%%%%%%%%%
%%%%%%%%%%%%%%%%%%%%%%%%%%%%%%%%%%%%%%%%%%%%%%%%%%%%%%%%%%%%%%%%%%%%%%%%%%%%%%%
% APPENDIX
%%%%%%%%%%%%%%%%%%%%%%%%%%%%%%%%%%%%%%%%%%%%%%%%%%%%%%%%%%%%%%%%%%%%%%%%%%%%%%%
%%%%%%%%%%%%%%%%%%%%%%%%%%%%%%%%%%%%%%%%%%%%%%%%%%%%%%%%%%%%%%%%%%%%%%%%%%%%%%%
\newpage
\appendix
\onecolumn

\section{Prompt} \label{appendix:1}
\begin{figure}[ht]
\vskip -0.1in
\begin{center}
\centerline{\includegraphics[width=0.87\textwidth]{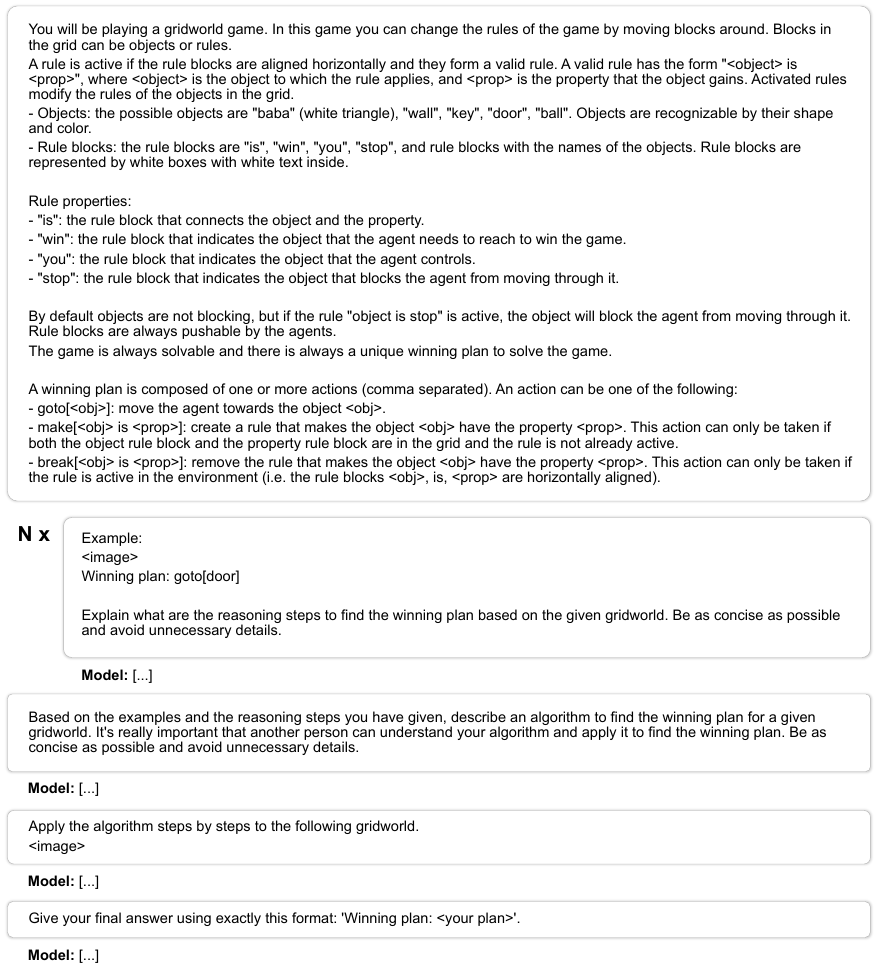}}
\caption{\textbf{Prompt template.} The first part of the prompt includes general instructions describing the game and the task for the LLMs. After providing the game instructions, we present LLMs with N = 10 example grid images and corresponding winning plans, and ask LLMs to generate reasoning steps to derive the plan from the grid image. We then ask LLMs to describe an algorithm to solve the environments and to apply to new environments that systematically differ from the in-context environments.}
\label{fig:insructions}
\end{center}
\vskip -0.4in
\end{figure}

\clearpage
\section{Error cases} \label{appendix:2}

\begin{figure}[ht]
\vskip 0in
\begin{center}
\centerline{\includegraphics[width=0.7\textwidth]{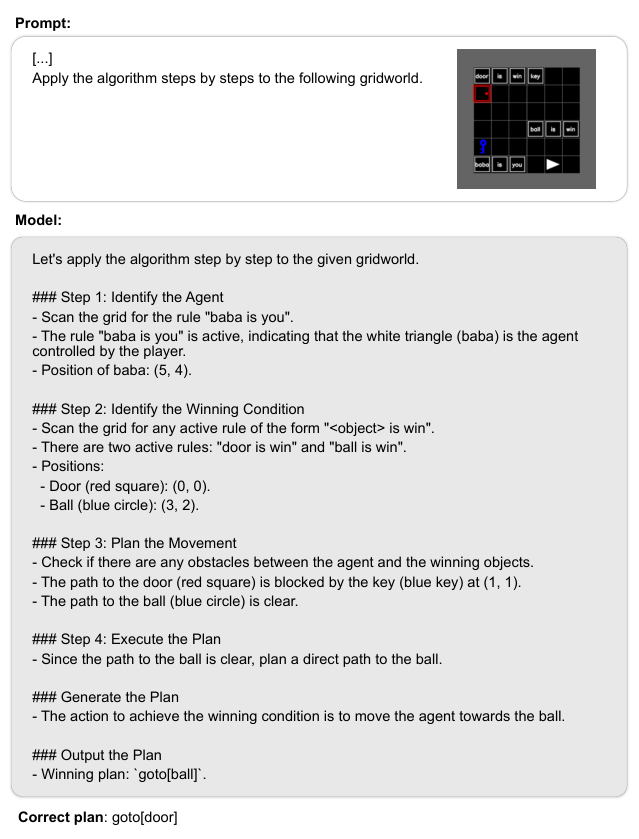}}
\caption{\textbf{Two common types of mistakes observed in the reasoning of LLMs}, illustrated here for GPT-4o when tested on the single-room environment with an additional win rule distractor. (i) Grounding mistake: the LLM refers to an object that does not exist in the environment. In this example, the LLM mentions a ball (Step 3, third bullet point in the model's answer), specifying that is it a blue circle in parenthesis, even though no such ball is present. (ii) Path planning mistake: the LLM incorrectly asserts that the path to a specific object is blocked by another object, despite the path being clear in the environment. In this instance, the LLM claims that the path to the door is blocked by the key, even though it is not the case.}
\label{fig:insructions}
\end{center}
\vskip -0.5in
\end{figure}

%%%%%%%%%%%%%%%%%%%%%%%%%%%%%%%%%%%%%%%%%%%%%%%%%%%%%%%%%%%%%%%%%%%%%%%%%%%%%%%
%%%%%%%%%%%%%%%%%%%%%%%%%%%%%%%%%%%%%%%%%%%%%%%%%%%%%%%%%%%%%%%%%%%%%%%%%%%%%%%

\end{document}